\documentclass[twocolumn,reqno]{article}

\usepackage[utf8]{inputenc}
\usepackage[T1]{fontenc}
\usepackage{newtxtext,newtxmath}
\usepackage[scaled=0.92]{helvet}

\usepackage{geometry}
\usepackage[mathlines,switch]{lineno}

\usepackage{amsmath}

\usepackage{amssymb}

\usepackage{amsfonts}

\usepackage{graphicx}
\usepackage{float}
\usepackage{placeins}
\usepackage{caption}
\usepackage{booktabs}
\usepackage{wrapfig}
\usepackage[table]{xcolor}
\usepackage{multirow}
\usepackage{makecell}

\usepackage{cite}
\usepackage{xurl}
\usepackage{hyperref}

\definecolor{bluegraypale}{RGB}{212,205,224}
\definecolor{bluegraywhisper}{RGB}{237,233,242}
\definecolor{bestred}{RGB}{200,0,0}

\begin{document}
\nolinenumbers

\twocolumn[
\begin{center}

{\huge\bfseries
A multimodal large language model for evidence-based autism spectrum disorder screening
\par}

\vspace{1.5em}

{\large
Jun Chen$^{1,2,4}$,
Qi Zhao$^{3,6}$,
Yunliang Jiang$^{1,2,7\dagger}$,
Shuqin Cao$^{1,5}$,
Yunqiang Lin$^{5}$,
Chenglong Jia$^{5}$,\\
Qiang Guo$^{5}$,
Guang Dai$^{6}$,
Xiongtao Zhang$^{7}$,
Mengmeng Wang$^{8}$,
Xiaoyue Ma$^{2}$, \\
Hui Wang$^{9}$,
and Wenguan Wang$^{10}$
\par}

\vspace{1em}

{
\small
$^{1}$National Special Education Resource Center for Children with Autism, Zhejiang Normal University, China\\
$^{2}$School of Computer Science and Technology, Zhejiang Normal University, China\\
$^{3}$School of Mathematics and Statistics, Xi'an Jiaotong University, China\\
$^{4}$Zhejiang Key Laboratory of Intelligent Education Technology and Application, Zhejiang Normal University, China\\
$^{5}$College of Child Development and Education, Zhejiang Normal University, China\\
$^{6}$SGIT AI Lab, State Grid Corporation of China, China\\
$^{7}$School of Artificial Intelligence, Huzhou Normal University, China\\
$^{8}$Zhejiang University of Technology, China\\
$^{9}$Children's Hospital, Zhejiang University School of Medicine, China\\
$^{10}$The State Key Lab of Brain-Machine Intelligence, Zhejiang University, China\\
$^{\dagger}$Correspondence should be addressed to:
jyl2022@zjnu.cn
}

\vspace{1em}

\begin{minipage}{0.9\textwidth}
\small

\textbf{Abstract:}
The clinical management of autism spectrum disorder (ASD) faces a bottleneck in early screening, mainly because trained specialists are scarce and conventional assessment tools are subjective. Here, we introduce ASDchat, a multimodal large language model designed for evidence-based ASD screening, which takes video, audio, and dialogue as input. ASDchat adopts a dual-branch architecture, where the decision branch generates screening probabilities and the evidence branch generates traceable, timestamped behavioral evidence aligned with the Autism Diagnostic Observation Schedule (ADOS-2). The model was trained and evaluated on a dataset of 1,035 participants from 27 sites in China, which covered typically developing (TD) children, children with ASD, and children with other developmental disorders. For ASD versus TD, ASDchat reached an area under the receiver operating characteristic curve (AUC) of 0.953 ± 0.021. On 6 held-out sites that were not used for training, the mean AUC was 0.961. Furthermore, unsupervised clustering of the behavioral dimensions split the ASD cases into six subtypes with different phenotypic profiles, and ASDchat suggests an intervention for each subtype. ASDchat provides a feasible path for large-scale, evidence-based ASD screening in clinical practice.

\vspace{1em}
\end{minipage}

\vspace{2.5em}
\end{center}
]

\section*{Introduction}

Autism spectrum disorder (ASD) is a neurodevelopmental disorder characterized by persistent differences in interactive social behaviors and restricted, repetitive behaviors. Multiple epidemiological monitoring results indicate that the prevalence of ASD worldwide is on the rise, and the clinical management of ASD has become a prominent challenge in the field of public health \cite{lyall2017changing,zeidan2022global}. The social and economic infrastructure required for long-term care is facing significant pressure \cite{lord2018autism,maenner2023prevalence}. Neurodevelopmental research indicates that the optimal intervention window for ASD is mostly before the age of 3, when the brain of children is highly plastic \cite{tang2014loss}. During this period, implementing targeted behavioral interventions is expected to reshape children's cognitive abilities and social adaptability \cite{marin2016developmental,dawson2010randomized}. Therefore, standardized developmental screening tools, especially the Modified Checklist for Autism in Toddlers (M-CHAT) \cite{robins2014validation}, have been widely used in primary care, helping to identify at-risk children early \cite{campbell2017use,zwaigenbaum2015early}. This demand is further magnified in centralized medical systems. For example, in China, a large number of at-risk children coexist with a concentrated group of specialized developmental pediatricians \cite{sun2019autism,zhou2020prevalence}. This has led to many families being on long waiting lists, delaying the children's access to care.

Clinical diagnosis and screening still rely mainly on the traditional assessment system. This system has limitations at both the operational and methodological levels. The current screening workflows are divided into two categories: caregiver-report questionnaires and clinician-administered structured observation. Questionnaire tools such as the M-CHAT and the Social Responsiveness Scale (SRS-2) \cite{bruni2014test} complete the assessment based on caregiver feedback. The assessment results are affected by factors such as parental education level, cultural stigma, and recall bias. This interference directly reduces the positive predictive value of questionnaire screening \cite{campbell2017use,constantino2021social}. The Childhood Autism Rating Scale (CARS-2) \cite{schopler2010childhood} and the Autism Diagnostic Observation Schedule (ADOS-2) \cite{lord2012autism} are the mainstream assessment tools used in clinical practice to avoid subjective errors. These two tools assess children based on standardized interactive tasks and ratings by certified specialists. Both tools have stable diagnostic validity, but their use in practice is limited. A complete assessment requires certified specialists and systematic training. The assessment process requires a large amount of human resources and time, and is not suitable for large-scale screening. Traditional assessment tools can only collect data on children's development at a single time point. Such cross-sectional data cannot reflect the long-term changes in children's behavioral phenotypes. While video recording can expand the data available for assessment, the manual coding and behavioral annotation work is time-consuming and subject to inter-rater variability. These conditions prevent the traditional observation mode from supporting standardized, automated large-scale screening \cite{wall2012use}.

To alleviate the workload of clinical personnel, researchers in the field of computer science have attempted to utilize machine learning and deep learning to develop an early automated screening approach for ASD \cite{insel2017digital,washington2023review}. These digital screening systems extract objective behavioral indicators from raw sensor data, which can replace manual assessment tools and are convenient for use in large populations. Most of the early related studies focused on analyzing localized behavioral characteristics. These studies utilized computer vision to extract abnormal features of two-dimensional skeletal postures \cite{kojovic2021using}, quantify gaze patterns and eye-contact avoidance behaviors \cite{chong2020detection,boluk2025gaze}, identify abnormal movement trajectories \cite{jabbar2026deep,khan2025ws}, or process resting-state functional magnetic resonance imaging data \cite{nafisah2025deep}. These automated models can complete classification tasks in controlled experimental environments. ASD is a multi-system neurodevelopmental disorder, and its behavioral phenotypes are heterogeneous and cross-modal \cite{lord2018autism}. A single-modality feature limits the model's ability to represent the heterogeneous nature of ASD. It may represent only part of this phenotype, and the model may fail to capture the correlation among facial expressions, motor responses, and linguistic feedback. In the face of the wide phenotypic variation among individuals with ASD, the model's accuracy, sensitivity, and specificity decrease.

In recent years, multimodal large language models (MLLMs) and foundation models have provided new directions for overcoming the limitations of single-modality methods \cite{moor2023foundation,singhal2023large,kim2025automated,kommineni2025can,yoo2025care}. These models typically adopt a joint vision-language architecture with various multimodal fusion modules, such as cross-attention, projection-based alignment, or adapter-based alignment, which can integrate different types of input data and understand the complex behaviors of children in interaction scenarios \cite{natraj2024video,deng2024hear,zhong2025multi}. Nevertheless, existing ASD screening models, even those built on MLLM architectures, still face obstacles in clinical application. Most of them follow a discriminative paradigm that directly converts high-dimensional feature vectors into probability outputs without an explicit evidence-based reasoning \cite{yoo2025care,kim2025automated}.  Pediatric screening is a high-risk scenario, consequently, a model that lacks a reasoning process conforming to clinical logic is difficult for clinicians and caregivers to trust \cite{marey2024explainability}. This lack of traceability makes the screening results difficult to verify and limits their utility for intervention.

To address the aforementioned issues, this paper proposes ASDchat, an evidence-based multimodal large language model for ASD screening. It incorporates structured interaction video sequences, quantitative acoustic data, and dialogue structure, and adopts a dual-branch design: a decision branch that estimates screening probability and an evidence branch that generates traceable, timestamped behavioral evidence aligned with the clinical criteria. The evidence branch links multi-channel behavioral observations, such as eye contact, joint attention deficits, and vocalization changes, to ADOS-2 items, ensuring that screening decisions are grounded in observable evidence. We train and evaluate ASDchat on a large multi-site dataset covering typically developing (TD) children, children with ASD, and children with other developmental disorders in China. This architecture allows the model to achieve robust screening performance and to generalize to held-out sites not used for training. Ablation studies further confirm that the evidence-based reasoning is essential to this performance, while audio and dialogue add a small gain over video. Beyond binary screening, unsupervised clustering of behavioral dimensions reveals six ASD subtypes, and ASDchat translates these subtypes into targeted intervention recommendations. Together, these results suggest that ASDchat offers a practical and interpretable path toward large-scale ASD screening in clinical practice.

\section*{Results}

\begin{figure*}[!htb]
    \centering
    \includegraphics[width=\textwidth]{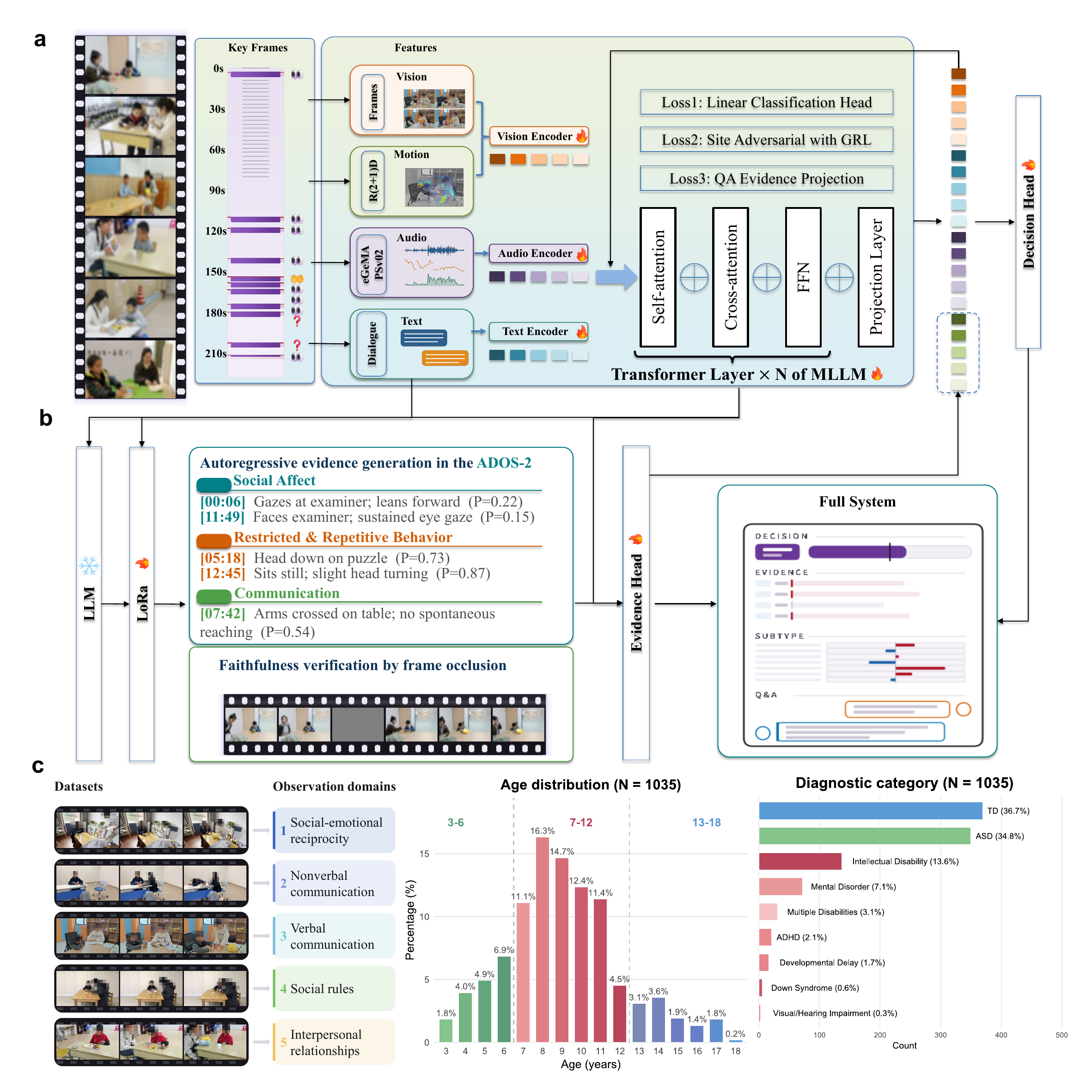}
    \caption{\textbf{Overview of ASDchat: a multimodal LLM for evidence-based question answering.} \textbf{(a)}, the decision branch of ASDchat is constructed based on visual, audio, and dialogue modalities; \textbf{(b)}, the evidence branch of ASDchat generates evidence chains based on ADOS-2, which carries a direction, a severity and a time window; \textbf{(c)}, dataset illustrates observation domains, showing distribution (age distribution and diagnostic category).}
    \label{fig_main1}
\end{figure*}

\subsection*{ASDchat Overview}

ASDchat takes a structured video recording of a child's interaction with an examiner as input. The system outputs the screening probability, the evidence chain with its time windows, and the responses to the spatio-temporal localization questions. Fig.~\ref{fig_main1} shows the overall system architecture. ASDchat consists of two branches: the multimodal decision branch for outputting probabilities, and the evidence branch for providing the basis for judgment.

The decision branch is built based on three complementary modalities, forming a multimodal fusion framework (Fig.~\ref{fig_main1}a). The visual modality extracts the child's behavior, the audio modality collects the child's speech, and the dialogue structure records the progression of the interaction. Each type of data is processed by a dedicated trainable encoder. The features output by the encoders are sent to the vision tower of the MLLM, and the screening probability is calculated.

\textbf{The visual path} is the main input. In the source video preprocessing stage, all pixels outside the contours of the people in the frame are masked to retain the contours of the child and the examiner, and all frames are converted to grayscale. The model extracts 64 frames from the complete recording based on social bids. Of the frame budget, up to 60\% is allocated to the three-second response window after each social bid, and the rest is sampled uniformly across the remaining recording. Temporal motion information is extracted in parallel by the spatio-temporal path, and six clips are selected, each containing 16 densely sampled frames at 112 $\times$ 112, which are processed by the frozen R(2+1)D model. The processed data is sent to the token encoder to generate visual modality features.

\textbf{The audio path} encodes the recording in windows centered on the key frames. This path extracts 88 extended Geneva Minimalistic Acoustic Parameter Set (eGeMAPS) descriptors, aggregates them into a 176-dimensional feature, and appends 28 interaction timing features. The model standardizes the features within each site, using statistics fitted on the training folds alone, and removes the site information before the features enter the fusion module. Similar to the visual path, the audio features are sent to the audio encoder, whose output enters the shared feature sequence as 64 tokens aligned with the key frames.

\textbf{The dialogue path} also encodes the recording in windows centered on the key frames. It extracts twelve behavioral statistics from the speaker-annotated text of the speech transcription, including the number of rounds, the number of statements initiated and responded by children, and the proportion of speaking time, without using any lexical content. Then, it is encoded by a text encoder and sent together with the other two modalities to the vision tower of the MLLM.

These contents are processed through the fine-tuned vision tower of the MLLM, then input into a bidirectional recurrent layer for aggregation, and finally generated into a single record vector $h_v$ through time averaging. This vector is then input into the decision head, the site discriminator and the evidence supervision head. The fine-tuning of this vision tower was conducted using a labeled multi-site dataset.

The evidence branch reads the same 64 frames through the low-rank adapter language model and generates evidence chains as well as question answers (Fig.~\ref{fig_main1}b). Each item carries a direction, a severity and a time window, so that a reader can go back to the recording and watch the segment the system is pointing at. The branch also answers questions in both directions, it names the moment at which a behavior occurs, and it states which behavior occurs at a moment the reader names. The faithfulness term keeps the probability near chance when all frames are replaced with uniform gray, so the decision cannot rest on a prior alone.

The two branches are interconnected, where the mean of the last-layer hidden states of the evidence branch is projected onto the $h_v$ dimension and concatenated with it. The resulting concatenated result is then passed through a fusion layer into the deployed decision head. Therefore, when the evidence representation enters the screening decision, it is no longer solely used to verify the description of the system, but rather makes decisions and presents evidence through visual reading. The above-mentioned effects are bidirectional. During the main training process, the evidence supervision loss is propagated in the opposite direction along with the classification loss, thus the evidence target will shape the aggregated representation before the connection of the language model. During deployment, the evidence representation will be re-injected before a new decision head.

These two branches finally form a complete verification system through collaborative cooperation. This system can not only output the probability of ASD but also provide clinical evidence, and give specific suggestions based on the recording. At the same time, parents and clinicians can both ask questions about specific segments of the video.

\subsection*{Dataset Description}

\begin{figure*}[!htb]
    \centering
    \includegraphics[width=\textwidth]{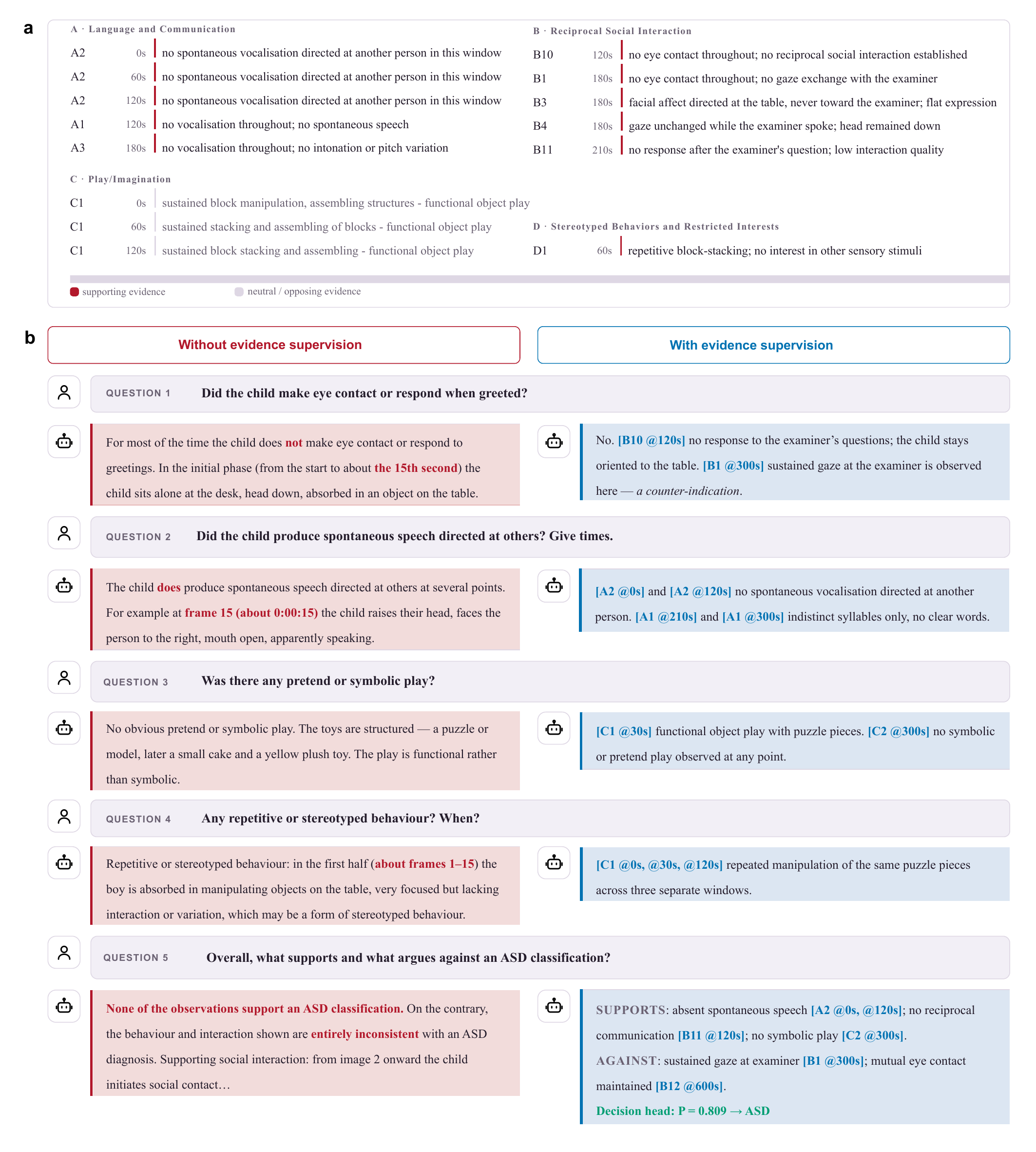}
    \caption{\textbf{ASDchat accurately screens for autism and provides video-based evidence.} \textbf{(a)}, video evidence exhibits behaviors matching four of the five ADOS-2 domains: Language and Communication, Reciprocal Social Interaction, Play/Imagination, and Stereotyped Behaviors and Restricted Interests; \textbf{(b)}, the multimodal model infers evidence both supporting and opposing the diagnosis, and provides the specific timestamps at which each relevant evidence occurs in the video.}
    \label{fig_main2}
\end{figure*}

\begin{figure*}[!htb]
    \centering
    \includegraphics[width=\textwidth]{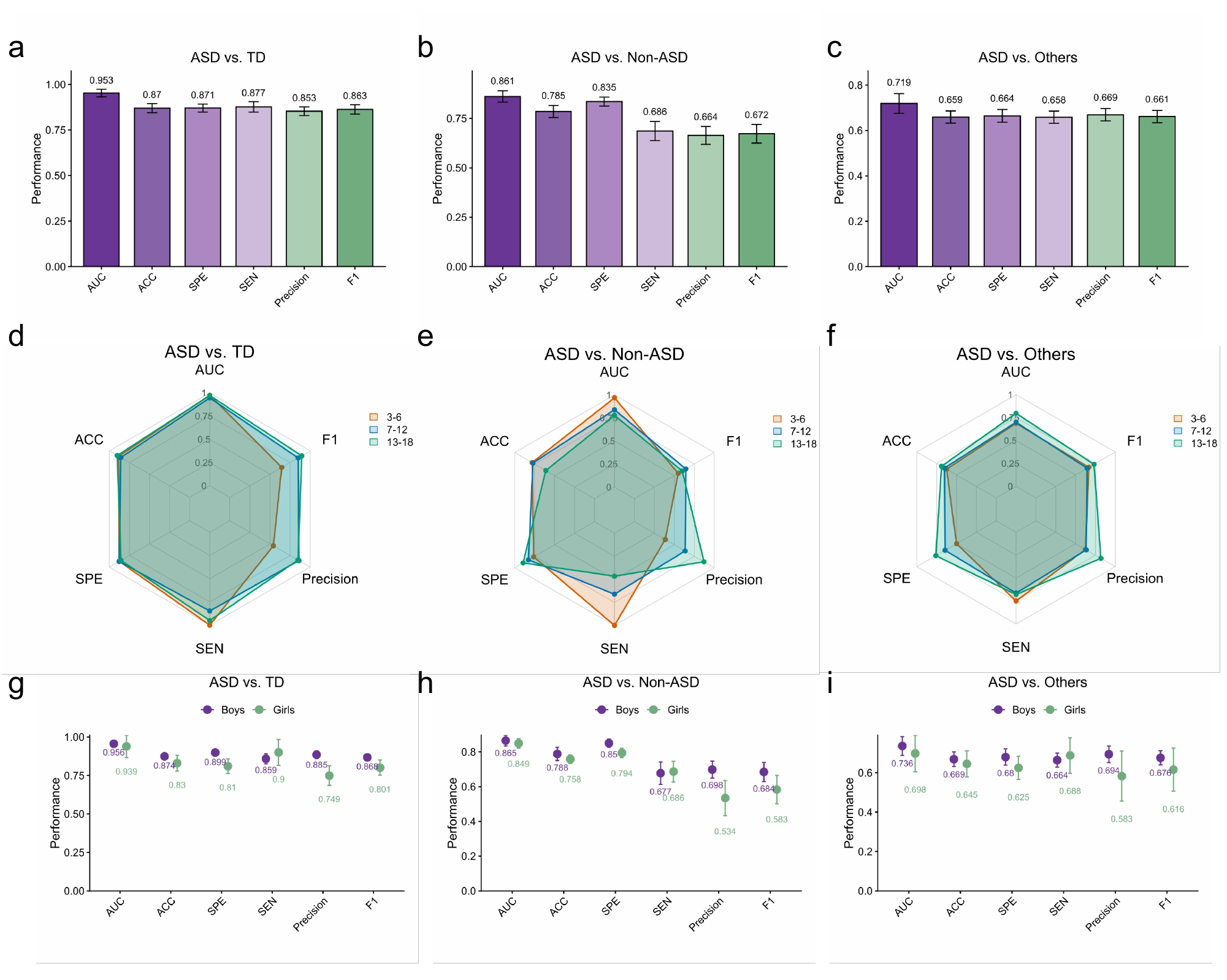}
    \caption{\textbf{Screening performance of ASDchat across classification tasks, age groups, and sex.} \textbf{(a-c)}, overall screening performance on three binary classification tasks: ASD vs. TD (excluding other developmental disorders), ASD vs. Non-ASD (including TD and other developmental disorders), and ASD vs. Others (other developmental disorders, excluding TD); \textbf{(d-f)}, performance stratified by age group (3-6, 7-12, and 13-18 years) for the same three tasks; \textbf{(g-i)}, performance stratified by sex (boys vs. girls) for the same three tasks. Metrics include AUC, ACC, SPE, SEN, Precision, and F1. All results are reported as Mean ± SD from 5-fold cross-validation. The three columns correspond to the three classification tasks, and the three rows correspond to overall, age-stratified, and sex-stratified analyses, respectively.}
    \label{fig1}
\end{figure*}

This dataset was jointly collected at multiple research sites across three provinces in China: Yunnan (9 sites), Hunan (5 sites), and Zhejiang (13 sites), covering 27 sites distributed across 9 districts. The dataset consists of 1,035 participants, including 370 typically developing children, 350 children diagnosed with ASD, and 315 children with other developmental disorders, such as Intellectual Disability, Mental Disorder, Attention-Deficit/Hyperactivity Disorder (ADHD), Developmental Delay, Visual/Hearing Impairment, Down Syndrome, and Multiple Disabilities (Fig.~\ref{fig_main1}c). Within the ASD group, there are 282 males and 68 females, with a male-to-female ratio of approximately 4:1, which is consistent with existing epidemiological estimates of ASD prevalence \cite{fyfe2026time}. The age range of the participants is 3 to 18 years, covering all developmental stages within this range. All participants or their legal guardians signed written informed consent before enrollment. The data collection plan was approved by the ethics committees of Zhejiang Normal University, the collaborating hospitals, and each participating site (No. ZSRT2026040).

Each participant engaged in a structured interaction of 5 to 10 minutes, and the session was video-recorded. This collection plan was designed to elicit natural behaviors across multiple developmental dimensions and to reduce the influence of external intervention on the children's responses. All sessions were conducted in environments familiar to the participants. A parent, caregiver, or teacher was present and interacted with the child naturally. The adults present received no specialized training or instructions, so that the recorded behaviors reflected social communication in real-life settings.

The dataset contains two types of core data:
\begin{itemize}
    \item Video with synchronized audio, which records the full interaction of each participant and preserves behavioral information such as facial expressions, body movements, gaze patterns, and vocalizations.
    \item Demographic and clinical information collected uniformly for each participant, including age, sex, educational stage, family income, parental education level, parental occupation, diagnostic category, diagnostic source, comorbid conditions, intellectual developmental level, and medical history.
\end{itemize}
This multimodal structure enables us to simultaneously observe behavioral signals and contextual factors, and to establish an integrated screening framework that incorporates both phenotypic and demographic information.

The session protocol consists of two types of tasks. The first is game-based tasks, such as tabletop games and a doll birthday celebration, which are used to engage the child in structured play. The second is interactive tasks, including construction play, picture description, and prompts for conversation about friendships and social experiences. These tasks correspond to the five behavioral dimensions examined in ASD screening:
\begin{enumerate}
    \item Social-emotional reciprocity: the child's ability to respond appropriately to task changes and social cues;
    \item Nonverbal communication: the frequency and quality of eye contact during social interaction;
    \item Verbal communication: the use of complex sentence structures in conversation;
    \item Social rules: the presence of off-task behaviors such as frequently leaving the seat or being distracted by other stimuli;
    \item Interpersonal relationships: the child's ability to adjust responses to changes in others' emotional states.
\end{enumerate}

All videos were recorded under standardized lighting and camera positions to ensure consistent collection conditions at each site. The samples were obtained from multiple research sites with different geographical and socioeconomic conditions in China. This gives the dataset broader applicability and makes it possible to compare ASD behavioral manifestations across regions. We did not simply divide the participants into ASD and typically developing groups. We also included a range of other developmental disorders. This makes the test of the screening algorithm's specificity more stringent, and it also avoids overfitting to the binary classification boundary. This dataset is suitable for training and evaluating multimodal machine learning models, especially large language models and vision-language models, for automated ASD screening and behavioral phenotyping.

\subsection*{ASDchat for Evidence-Based Reasoning}

\begin{figure*}[!h]
    \centering
    \includegraphics[width=\textwidth]{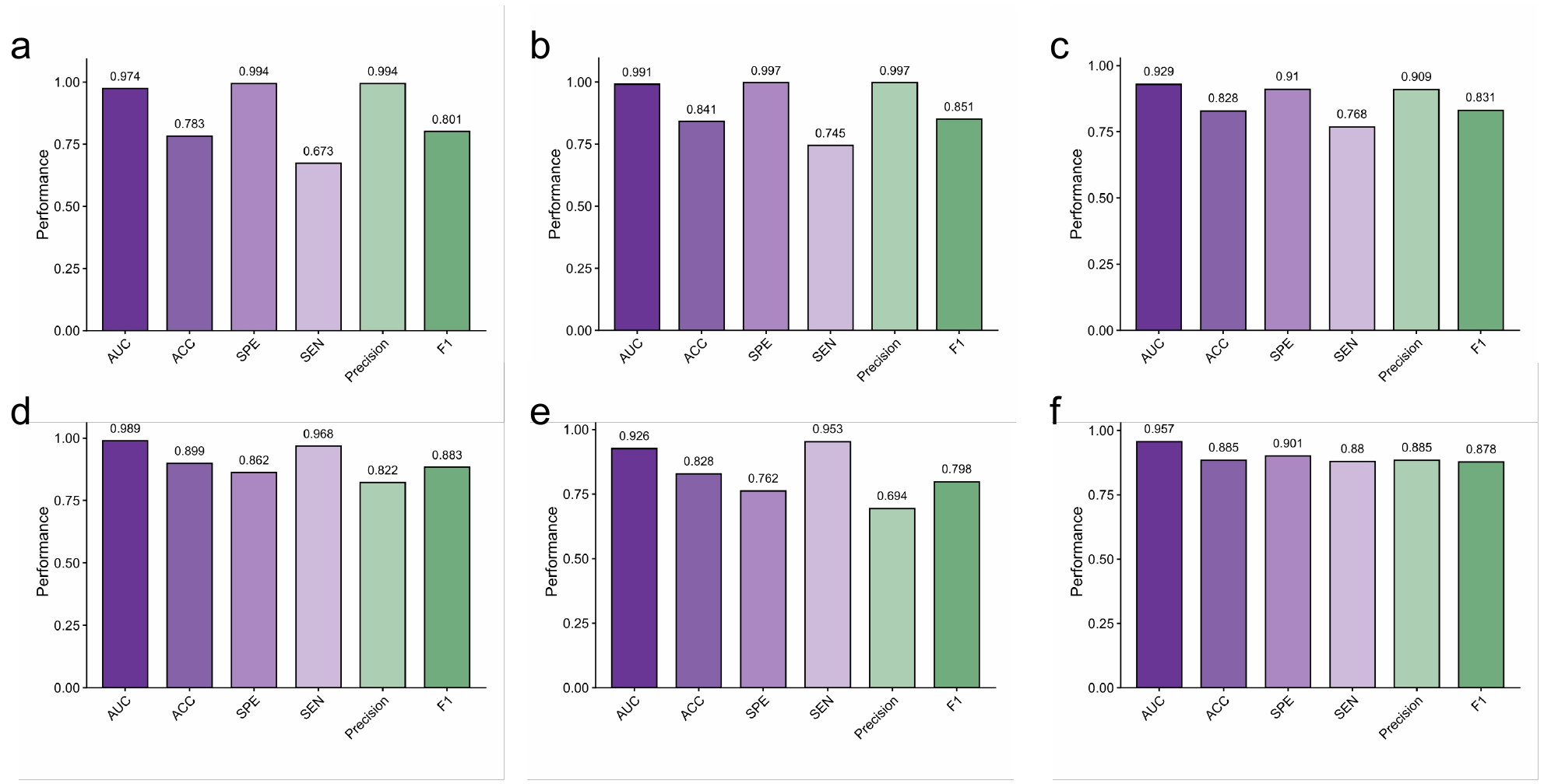}
    \caption{\textbf{Cross-site Performance of ASDchat.} \textbf{(a-f)}, six combined held-out sites were formed from sites randomly selected among the 27 sites for cross-site screening, ensuring that each site includes both typically developing children and children with ASD for model evaluation.}
    \label{fig5}
\end{figure*}

\begin{figure}[h]
    \centering
    \includegraphics[width=0.5\textwidth]{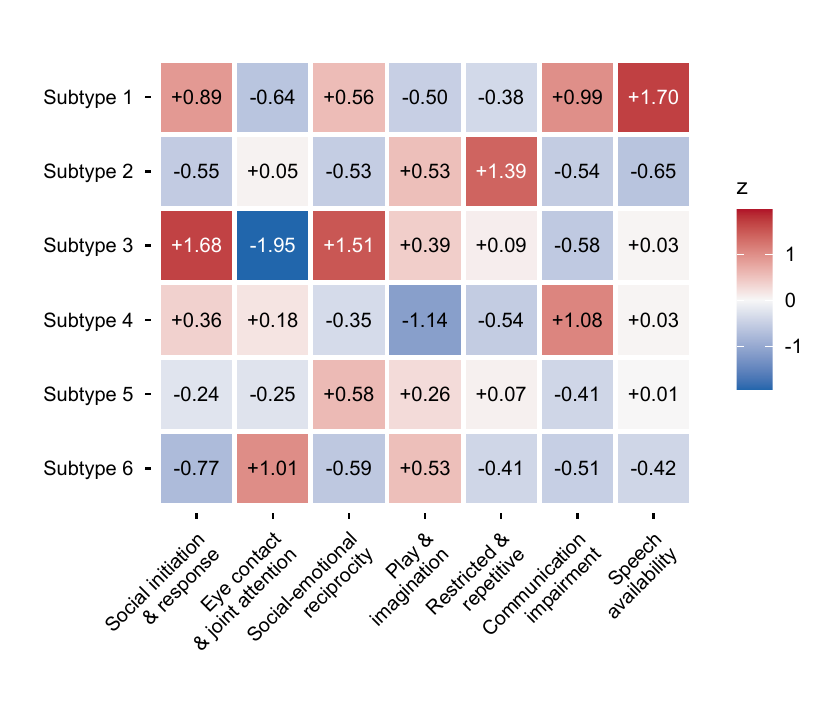}
    \caption{\textbf{Six behavioral subtypes of autism spectrum disorder using K-means clustering.}}
    \label{fig4}
\end{figure}

\begin{figure*}[!htb]
    \centering
    \includegraphics[width=\textwidth]{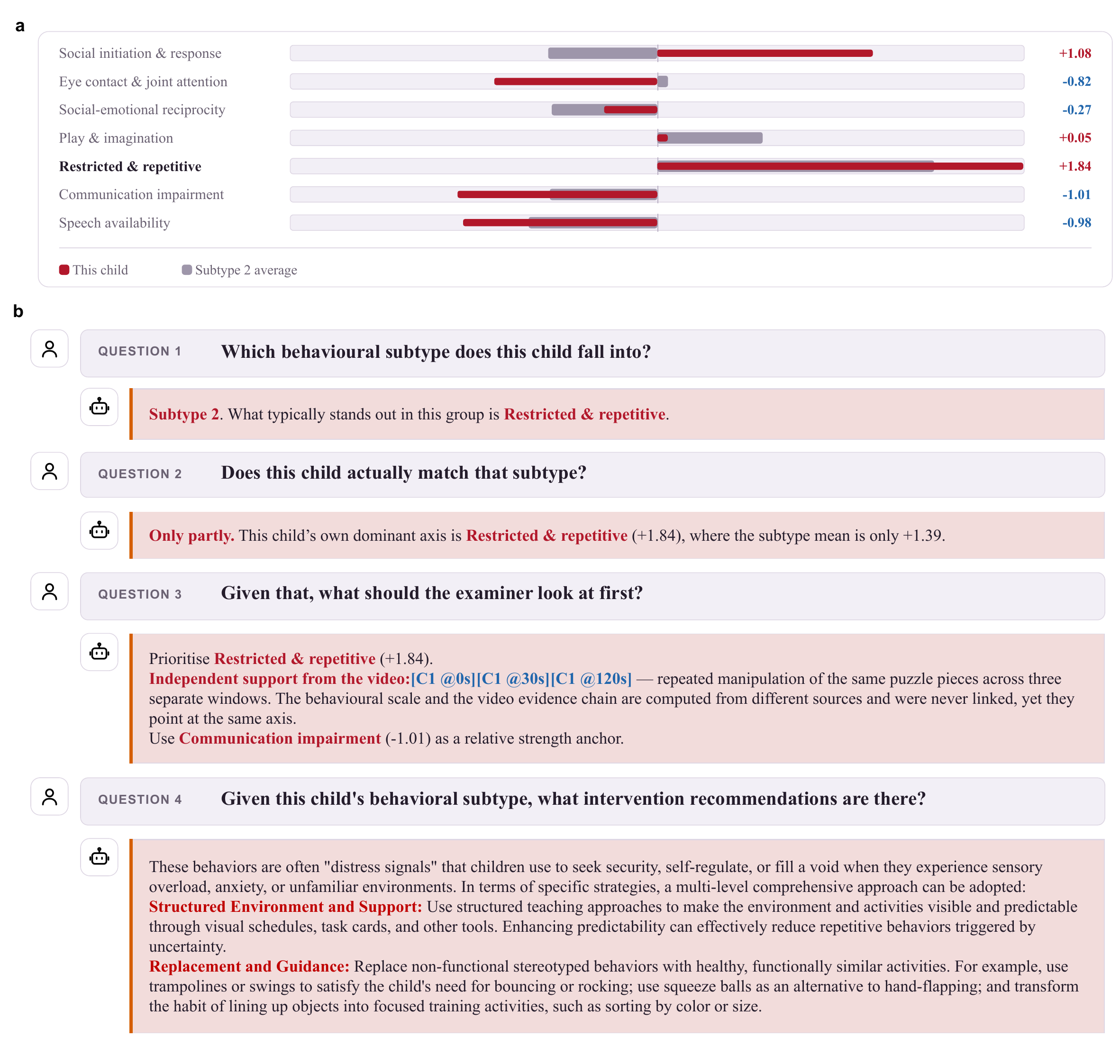}
    \caption{\textbf{ASDchat has the ability to classify subtypes of autism spectrum disorder.} \textbf{(a)}, an example of autism, illustrating its contrast with Subtype 2 across seven dimensions; \textbf{(b)}, the multimodal model determines the autism subtype for the case, along with supporting evidence and suggested interventions.}
    \label{fig_main3}
\end{figure*}

Fig.~\ref{fig_main2} illustrates the process of ASDchat conducting evidence-based screening, showing how it uses multimodal input to perform transparent and clinically interpretable reasoning. We use a vision-language architecture consistent with the standard diagnostic framework to assign the observed child behaviors to the five domains of the ADOS-2: Language and Communication, Reciprocal Social Interaction, Play/Imagination, Stereotyped Behaviors and Restricted Interests, and Other Abnormal Behaviors, which can be further broken down into finer sub-domain indicators (Fig.~\ref{fig_main2}a). The complete list of items, adapted from the ADOS-2 and used as annotation categories rather than scored with the ADOS-2 algorithm, is provided in Supplementary Table 1. This gives the model a temporal memory that records the exact moment each behavioral event occurs, so that evidence can be retrieved and each screening judgment has a basis. The system also adds a verification step to prevent the model from producing unsupported or fabricated evidence. The verification checks whether the cited ADOS-2 item matches the corresponding video segment, and confirms that the behavior seen in that segment is consistent with the judgment. Only judgments that pass both checks are retained. The rest are flagged or removed. ASDchat's judgments therefore stay within the behaviors that the ADOS-2 can observe, and each output can be traced back to clinical evidence.

Fig.~\ref{fig_main2}b contrasts two reasoning methods: without evidence supervision (left panel) and with evidence supervision (right panel). The model trained without evidence supervision can still output text, but this text cannot be used, so only its probability score or coarse binary classification is useful, and it provides neither a usable reasoning process nor usable evidence. Such results are difficult to trust in high-risk clinical scenarios and cannot provide a verifiable basis for subsequent intervention plans. ASDchat does not follow this approach, which identifies both supporting and opposing evidence for each screening question, and labels each piece of evidence with an exact video timestamp. The right panel provides an example: for the question ``Did the child look at the examiner and respond when greeted'', the model records that at about 120 seconds the child did not respond to the examiner and remained oriented to the table ([B10 @120s]), which it takes as evidence supporting an ASD classification. At about 300 seconds, the model records that the child maintained gaze toward the examiner ([B1 @300s]), which points in the opposite direction. Such conflicting cues are usually missed by models trained without evidence supervision, but ASDchat displays them explicitly. For spontaneous speech, symbolic play, and repetitive behaviors, the model also provides the corresponding temporal localizations and evidence descriptions, and all of this information can be verified. The model finally outputs a screening probability. Because every judgment is tied to a timestamped video segment, the screening result can be checked against the recording. Clinicians can view and verify the behavioral evidence behind each screening conclusion, which makes the system easier to accept in practice.

\subsection*{Comprehensive Quantitative Analysis}

\begin{figure*}[h]
    \centering
    \includegraphics[width=\textwidth]{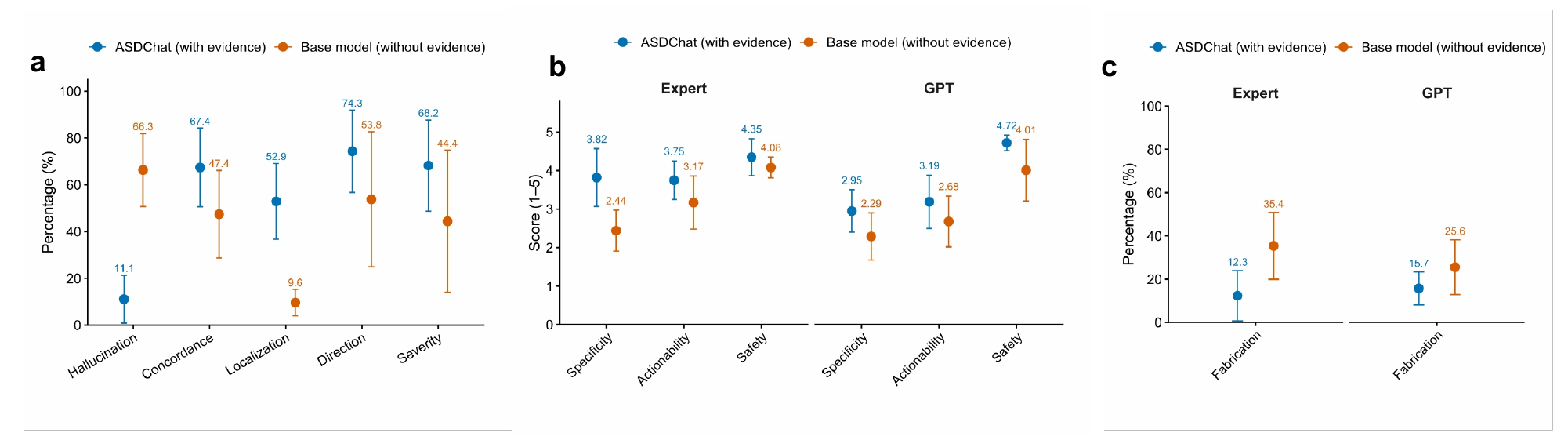}
    \caption{\textbf{Evaluation of the screening and suggestions.} \textbf{(a)}, screening quality with and without evidence. Metrics include Hallucination, Concordance, Localization, Direction, and Severity (Hallucination: lower is better; all other metrics: higher is better); \textbf{(b-c)}, the quality of intervention recommendations with and without evidence. Metrics include Specificity, Actionability, Safety, and Fabrication (Fabrication: lower is better; all other metrics: higher is better).}
    \label{fig7}
\end{figure*}

\begin{figure*}[h]
    \centering
    \includegraphics[width=\textwidth]{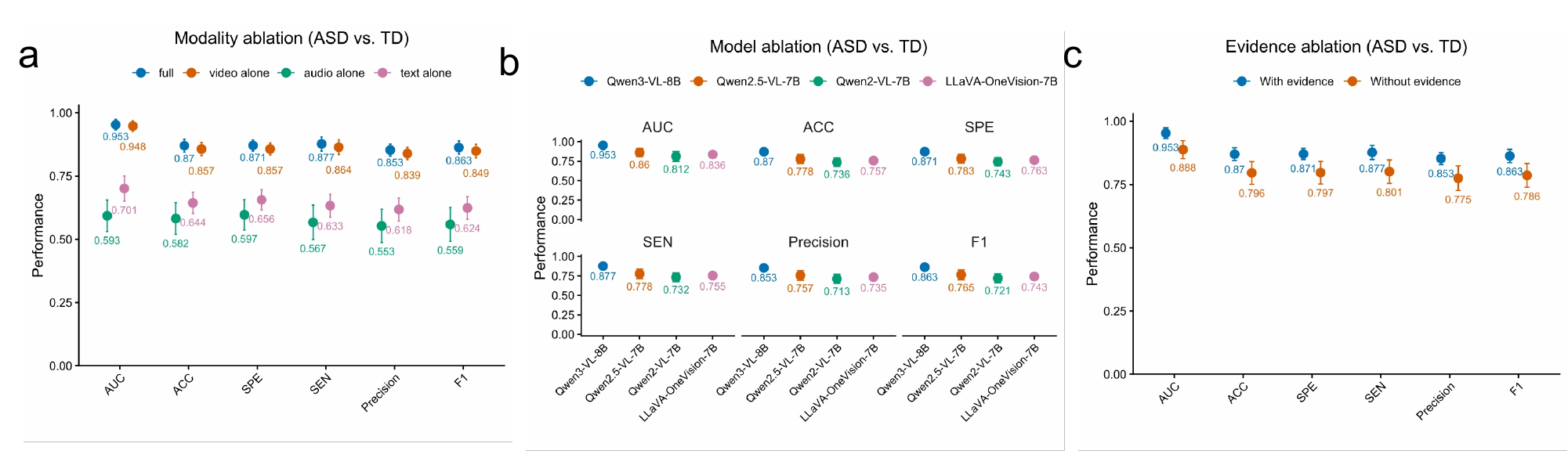}
    \caption{\textbf{Ablation studies of ASDchat evaluated on ASD and TD cases.} \textbf{(a)}, modality ablation for video, audio, and text; \textbf{(b)}, model ablation for Qwen3-VL-8B, Qwen2.5-VL-7B, Qwen2-VL-7B, and LLaVA-OneVision-7B; \textbf{(c)}, evidence ablation with and without evidence-based reasoning.}
    \label{fig6}
\end{figure*}

We evaluated the screening performance of ASDchat by setting up three binary classification tasks: ASD versus TD (excluding other developmental disorders), ASD versus non-ASD (including TD and other developmental disorders), and ASD versus other developmental disorders (excluding TD). All metrics were calculated through 5-fold cross-validation and the results are presented as Mean ± SD (Fig.~\ref{fig1}). On three binary classification tasks (Fig.~\ref{fig1}a-c), the model's AUC reached 0.953 ± 0.021 when distinguishing ASD from TD, with balanced sensitivity and specificity. The classification results for ASD versus non-ASD decreased, with an AUC of 0.861 ± 0.029. The task comparing ASD with other developmental disorders was the most difficult, with an AUC of 0.719 ± 0.043. The drop in this group is most likely due to the high comorbidity rate between ASD and other developmental disorders, and to their overlapping behavioral characteristics, which are common difficulties in clinical diagnosis. These results indicate that ASDchat can effectively distinguish ASD from TD children, but there is still considerable difficulty in distinguishing ASD from other developmental disorders with similar symptoms.

We grouped the samples by age to observe how model performance changes with developmental stage, and divided them into three groups: 3-6 years, 7-12 years, and 13-18 years (Fig.~\ref{fig1}d-f). In the ASD versus TD task, the youngest group (0.975 ± 0.032) and the oldest group (0.972 ± 0.016) had the highest AUC. The 3-6 group reached perfect sensitivity, but its precision and F1 score were much lower, because ASD cases make up only a small share of this age group, so a few false positives are enough to lower the precision. In the ASD versus non-ASD task, the 3-6 group performed best, with an AUC of 0.968 ± 0.064. The AUC decreased to 0.778 ± 0.147 in the 13-18 group, with larger variance. Among all age groups, distinguishing ASD from other developmental disorders was the most difficult, with AUCs below 0.8 and large standard deviations. These results indicate that the cognitive and behavioral characteristics associated with different ages affect the model's accuracy. In some developmental stages, the behavioral phenotypes shared by ASD and other developmental disorders are difficult to separate.

Fig.~\ref{fig1}g-i analyzes the differences in model performance between sexes. Overall, the model performed slightly better on boys than on girls in all tasks, but the ranges of mean $\pm$ standard deviation overlapped and the difference between the two groups was small. In the ASD versus TD task, boys reached an AUC of 0.956 ± 0.015, compared with 0.939 ± 0.073 for girls. In the ASD versus non-ASD task, the gap between boys and girls was clearer, with higher precision and F1 score for boys. Part of the reason is the unbalanced sex distribution in the dataset, where the male-to-female ratio is approximately 4:1. The small number of girls leads to larger variance, and also makes it harder for the model to learn behavioral patterns specific to girls, which lowers the average performance and increases its variability. This suggests that a dataset with a more balanced sex distribution and sex-stratified evaluation are needed to ensure stable and reliable screening across populations.

In summary, ASDchat identifies ASD and TD children stably, but model performance is affected by age and sex. When the comparison group shares clinical features with ASD, as in the non-ASD and other-disorder tasks, model performance drops markedly. This shows that distinguishing ASD from other developmental disorders is inherently difficult, which is consistent with the uncertainty encountered in clinical diagnosis. The results also indicate that stratified performance reporting and balanced sample collection can help improve the fairness and generalizability of the model.

\subsection*{Cross-site Performance Analysis}

We evaluated the generalizability of ASDchat across different sites. In the setting, we randomly selected sites from the 27 collected sites and combined them into 6 test sites (Fig.~\ref{fig5}). This combination strategy was adopted to prevent the situation where some sites contain only typically developing children or only children with ASD, which would make it impossible to evaluate the model. On this portion of the data, which was not involved in the training, we tested the screening performance of the model in distinguishing ASD from TD children. The mean AUC of the model on these 6 test sites was 0.961 ± 0.029, and the mean accuracy was 0.844 ± 0.042. However, the results varied across sites. The lowest AUC was 0.926 (Site e) and the highest was 0.991 (Site b), while accuracy ranged from 0.783 to 0.899. Site d had the highest sensitivity, reaching 0.968. Site e had a precision of only 0.694 and an F1 score of 0.798, which shows that in some deployment settings it is hard to reduce false positives and false negatives at the same time.

The performance varies among the sites, mainly because of differences in sample composition and inclusion criteria across the research sites. Such distribution shift is common in multi-center clinical studies and affects the stability of machine-learning-based screening tools \cite{fortin2018harmonization}. Despite these site differences, the overall metrics still have clinical value. When ASDchat is applied to data from a new site without site-specific fine-tuning, it can still perform the classification.
Overall, ASDchat generalizes acceptably across different clinical environments. However, the performance variation across sites must be considered before practical deployment.

\subsection*{Subtypes of ASD Behaviors}

Subtyping at the neuroimaging (fMRI) \cite{pagani2026autism,shnitko2026parsing}, molecular \cite{buch2023molecular}, and genomic \cite{litman2025decomposition} levels has deepened our understanding of the neurobiological heterogeneity of ASD. However, these methods have high implementation costs and limited scalability, and they are not closely related to the behaviors observable in daily life, which makes them difficult to apply directly in clinical settings. Subtyping at the behavioral level is more practical. It requires no specialized equipment, is non-invasive and readily scalable, and is suitable for early identification and personalized intervention. However, no empirically grounded behavioral classification system has yet been established for ASD.

To address this, we performed unsupervised K-means clustering on 350 children diagnosed with ASD. The input for each child was a seven-dimensional vector computed statistically from the per-window annotations produced by Qwen3-VL on the structured video recordings and reviewed by an expert, rather than from the video frames or the features of ASDchat: Social initiation \& response, Eye contact \& joint attention, Social-emotional reciprocity, Play \& imagination, Restricted \& repetitive behaviors, Communication impairment, and Speech availability. The number of clusters was determined by the silhouette coefficient \cite{rousseeuw1987silhouettes}, which selected six clusters. Fig.~\ref{fig4} presents the phenotypic profiles of the six subtypes across the seven dimensions. The six subtypes are clearly separated and clinically interpretable.
To illustrate how this subtyping can be used in clinical practice, Fig.~\ref{fig_main3} shows the multimodal framework applied to an actual case. For a case assigned to Subtype 2, the model compares the behavioral scores of the individual across the seven dimensions with the average scores of Subtype 2 on each dimension (Fig.~\ref{fig_main3}a). It identifies Restricted \& repetitive behaviors and Communication impairment as the dominant deficits, and on this basis assigns the subtype (Fig.~\ref{fig_main3}b). ASDchat then recommends personalized intervention plans, including targeted social-emotional training and structured communication support, which turns the subtype classification into practical clinical guidance.
Overall, the results indicate that ASD includes at least six reproducible, behaviorally defined subtypes. This classification system can help us screen earlier and more accurately, and can also support more targeted social-communication interventions. This behavior-based subtyping further contributes to a multi-level understanding of ASD heterogeneity.

\subsection*{Evaluation}

We further evaluated the quality of the evidence-based screening and intervention recommendations on 100 held-out recordings (Fig.~\ref{fig7}), comparing ASDchat with a base model trained without evidence supervision.

For the evidence-based screening (Fig.~\ref{fig7}a), the base model's grounding ability is poor. Its hallucination rate is nearly 6 times that of ASDchat (\(66.3 \pm 15.6\%\) compared to \(11.1 \pm 10.2\%\)), and the temporal localization is close to a random level (\(9.6 \pm 5.6\%\) compared to \(52.9 \pm 16.2\%\)). Additionally, ASDchat's performance in terms of concordance, direction, and severity has consistently improved. This indicates that the base model associates the overall video representation with the diagnostic labels, but fails to bind these labels with specific behavioral events. The evidence supervision requires the shared representation to encode the social evidence of the recording before classification. For evidence-based reasoning, our model must make judgments based on timestamped behavioral evidence. This simultaneously improves the accuracy of localization.

For the intervention suggestions (Fig.~\ref{fig7}b-c), 3 experts, including developmental behavioral pediatricians and special education teachers, scored the suggestions together with GPT-6 (gpt-6-astra, accessed on 7 October 2026) in a double-blind randomized comparison. On all dimensions, ASDchat was significantly superior to the base model. The greatest improvement in expert scores was in specificity (\(3.82 \pm 0.75\) compared to \(2.44 \pm 0.53\)), followed by actionability and safety, while fabrication dropped from \(35.37 \pm 15.48\%\) to \(12.32 \pm 11.66\%\). The GPT-6 scores also showed the same trend. The intervention suggestions rely on both the evidence chain and the assigned behavioral subtypes. The evidence chain provides behavioral deficits related to time for specific cases, while the subtypes provide normative phenotypic features. This dual conditioning avoids language models giving general recommendations, as they provide more personalized recommendations that are more in line with a child's main deficits. The reduction in fabrication and the improvement in safety are likely due to the same mechanism. When the intervention must be consistent with traceable evidence, recommendations that lack support or may be unsafe are less likely to be retained. The consistency between expert and GPT-6 scores further indicates that these improvements are also clearly visible in the intervention itself.

\subsection*{Ablation Study}

We analyzed the impact of each design choice in ASDchat one by one, with all ablations evaluated on the ASD versus TD task. Fig.~\ref{fig6} shows three types of ablation: input modality, vision-language backbone, and evidence integration.

We first examined the effect of modality (Fig.~\ref{fig6}a). The full multimodal model achieved an AUC of 0.953 ± 0.021. When only video was retained, the result was similar, with an AUC of 0.948 ± 0.021. This indicates that visual behavioral features are the dominant predictive signal. When using only audio, the AUC dropped to 0.593 ± 0.062, and when using only text, it was 0.701 ± 0.050. As independent information sources, both have limited discriminative ability. When the three modalities were combined, all metrics exceeded their respective single-modality versions, which indicates that audio and text refine the predictions from video, although the gain is smaller than the fold-to-fold variation.

We next compared different vision-language backbones (Fig.~\ref{fig6}b). The deployed Qwen3-VL-8B has an AUC of 0.953 ± 0.021, markedly higher than Qwen2.5-VL-7B (0.860 ± 0.053), Qwen2-VL-7B (0.812 ± 0.063), and LLaVA-OneVision-7B (0.836 ± 0.036). All metrics show stable improvements, which indicates that the architecture and training methods of newer vision-language models produce more reliable multimodal representations for ASD screening. The results of Qwen3-VL-8B also fluctuate less, with an AUC standard deviation of 0.021, while the earlier Qwen versions range from 0.053 to 0.063. Predictions are more stable across data splits, which matches what clinical deployment requires.

We finally isolated the role of the evidence branch (Fig.~\ref{fig6}c). The full model with this branch achieved an AUC of 0.953 ± 0.021, while removing it dropped the AUC to 0.888 ± 0.036. Performance declined markedly. Accuracy, sensitivity, and F1 score all dropped, which indicates that this branch provides complementary information beyond the raw behavioral signals. The branch also ties the predictions to ADOS-2 video evidence, which reduces fabricated evidence and makes the screening results more reliable in clinical settings. The evidence branch is therefore an essential component that grounds the model's decisions in observable behavioral markers.

\section*{Methods}




\subsection*{Video Preprocessing}

The preprocessing is carried out at the source video stage rather than frame-by-frame. The Mask R-CNN instance segmentation \cite{he2017maskrcnn} is applied to the video, and the union of all human masks is taken. All the external pixels in this union are set to zero. The remaining part is converted into a grayscale image. This union includes the examiner and the child, so this operation can separate the people from the video scene, but it cannot distinguish between the child and the examiner. In the actual deployment input, the average proportion of non-zero pixels per frame is 22.5\%.

\subsection*{Key Frame Selection}

The examiner's speech is labeled as the speaker label \cite{bredin2020pyannote,gao2023funasr}, and social bids are detected from the examiner utterances \cite{lord2012ados2,constantino2021social}. For the social bid ending at time $t_i$, the response window is $W_i=[t_i, \, t_i+\tau]$, where $\tau=3\,\mathrm{s}$. Given a budget of $M=64$ frames, at most a proportion $\rho=0.60$ is allocated to the response windows and distributed among the $B$ windows in proportion to the number of candidate frames in each window,

\begin{equation}
\begin{gathered}
m_i=\max\Big(1,\ \mathrm{round}\Big(M_r \cdot \frac{n_i}{\sum_{j=1}^{B} n_j}\Big)\Big),\\
M_r=\min\big(\mathrm{round}(\rho M),\ N_r\big),
\end{gathered}
\end{equation}

where $n_i$ is the number of candidate frames in $W_i$ and $N_r$ is the number of candidate frames that fall in any response window. Because every window lasts $\tau$, the split is close to equal. The frames selected from all windows are merged and duplicates are removed, and if overlapping windows push the count above $M_r$, the earliest frames are kept. The remaining frames are distributed uniformly throughout the video, and if the total exceeds $M$, the set is uniformly subsampled to $M$ frames. When no social bid is detected, all $M$ frames are sampled uniformly.
Uniform sampling ensures that recordings with few detected bids are still covered. The frames are sampled at a resolution of $256\times144$.

\subsection*{Audio, Dialogue and Motion Features}

In each analysis window, the audio channel extracts 88 eGeMAPS descriptors \cite{eyben2016gemaps}. These descriptors are aggregated by calculating the mean and standard deviation to generate a 176-dimensional feature. Additionally, 28 interaction timing features are appended.
The site effect is then removed by standardizing within each site, with the statistics fitted on the training folds alone,
\begin{equation}
\tilde{a}_{i}=\frac{a_{i}-\mu_{s(i)}}{\sigma_{s(i)}},\qquad
\mu_{s},\sigma_{s}\ \text{estimated on}\ \{i \in \mathrm{train},\ s(i) = s\}.
\end{equation}
The same standardization is applied to the dialogue and motion features. Recordings from sites absent from the training folds, including all held-out sites, and from sites with fewer than five training recordings are standardized with the pooled training statistics.
The dialogue channel extracts 12 behavioral statistics from the transcribed text with speaker labels, including the number of rounds, the number of statements initiated and responded by children, and the proportion of speaking time. No lexical content is used. The motion channel, which belongs to the visual modality together with the frames, samples six clips of 16 frames at stride two and resolution $112\times112$, and inputs these clips into an R(2+1)D-18 network \cite{tran2018r21d} pretrained on Kinetics-400 \cite{kay2017kinetics}, which is kept frozen. In each analysis window, the 512-dimensional features of the six clips are averaged, and no label is used at this stage. The analysis windows of the three channels are centered on the 64 key frames, so each channel yields 64 tokens aligned with the frame sequence.

\subsection*{Encoder and Aggregation}
The 64 frames are input into the vision tower of the MLLM (Qwen3-VL-8B \cite{qwen3vl2025}), which is fully fine-tuned, and the audio, dialogue, and motion tokens are fed into the vision tower together with the frame patches. For each of the other three channels $m$, the site-standardized token $\tilde{a}^{(m)}_t \in \mathbb{R}^{d_m}$ at key frame $t$ is mapped into the input space of the vision tower,
\begin{equation}
z^{(m)}_t = \mathrm{LN}\big(W_m\,\tilde{a}^{(m)}_t + b_m\big) \in \mathbb{R}^{d_0}, \qquad m \in \{\mathrm{aud},\mathrm{dia},\mathrm{mot}\},
\end{equation}
where $\mathrm{LN}$ is layer normalization, $d_0=1152$ is the width of the vision tower, and $W_m$ and $b_m$ are trained together with the vision tower. The projected tokens of the three channels are appended to the patch tokens $p_t$ of frame $t$,
\begin{equation}
\tilde{p}_t = \big[\,p_t;\ z^{(\mathrm{aud})}_t;\ z^{(\mathrm{dia})}_t;\ z^{(\mathrm{mot})}_t\,\big],
\end{equation}
so the frame, audio, dialogue and motion information at the same key frame is read by the vision tower in one sequence. A channel that is missing for a recording, or dropped during training, is omitted from the sequence. The output embedding values within each frame are averaged to obtain a sequence $x_{1:T} \in \mathbb{R}^{T\times d}$, where $T=64$ and $d=4096$. A bidirectional gated recurrent unit \cite{cho2014gru} with a hidden dimension of 128 reads this sequence and the representation is its mean output over time,
\begin{equation}
h_v = \frac{1}{T}\sum_{t=1}^{T} \mathrm{BiGRU}(x_{1:T})_t \in \mathbb{R}^{256}.
\end{equation}
Because $h_v$ is computed from the fused sequence, the heads below all act on a joint representation of the four channels. After $h_v$, three heads are appended, a decision head $s_{\mathrm{asd}}$, a site discriminator $s_{\mathrm{site}}$, and a question answering (QA) projection head $s_{\mathrm{qa}}$ that maps to a 200-dimensional evidence target.

\subsection*{Training Objective}

The site discriminator predicts whether a recording comes from the largest recruitment site or from any other site, and it is located after the gradient reversal layer (GRL) $R_\alpha$ \cite{ganin2015grl}. This layer remains identity during forward propagation, but in the backward propagation, it inverts and scales the gradients,
\begin{equation}
R_\alpha (u) = u, \qquad \frac{\partial R_\alpha }{\partial u} = -\alpha I.
\end{equation}
The coefficients follow the conventional pattern of the training progress, with a value of $p$ ranging from 0 to 1,
\begin{equation}
\alpha(p) = \lambda_{\mathrm{grl}}\left(\frac{2}{1+e^{-10p}} - 1\right)
\end{equation}
This keeps the reversed gradient weak early in training, so that the site discriminator is fitted before the reversal reaches full strength. The evidence supervision is a cosine term that acts on the evidence target $q$. For each recording, the silver-standard answers, reviewed by an expert, to three social questions, on the child's response, on joint attention and eye contact, and on the child's vocalization, are concatenated, encoded with character 2--4-gram TF-IDF, and reduced to 200 dimensions with truncated singular value decomposition, both fitted on the training folds only. The faithfulness term is a penalty on the logits produced for uniform gray frames $\bar{x}$, where each frame is replaced with a constant image of pixel value $(128,128,128)$, and the sequence maintains its length,
\begin{equation}
\mathcal{L}_{\mathrm{qa}} = 1 - \cos\big(s_{\mathrm{qa}}(h_v),\,q\big),
\qquad
\mathcal{L}_{\mathrm{faith}} = \left\|s_{\mathrm{asd}}(h_v(\bar{x}))\right\|_2^2.
\end{equation}
The second term causes the logit of the uniform gray frames to approach zero, meaning their probability approaches 0.5. Each channel also has an auxiliary head, a linear layer on its time-averaged tokens, and $\mathcal{L}_{\mathrm{aux}}$ is the mean binary cross-entropy of these heads. During training, each channel is dropped as a whole with a probability of 0.3, so the decision head cannot rely on any single channel. The complete objective function is,
\begin{equation}
\mathcal{L} = \mathcal{L}_{\mathrm{cls}}
+ \mathcal{L}_{\mathrm{site}}
+ \lambda_{\mathrm{qa}}\mathcal{L}_{\mathrm{qa}}
+ \lambda_{\mathrm{faith}}\mathcal{L}_{\mathrm{faith}}
+ \lambda_{\mathrm{aux}}\mathcal{L}_{\mathrm{aux}},
\end{equation}
Here, $\lambda_{\mathrm{grl}} = 1.0$, $\lambda_{\mathrm{qa}} = 0.5$, $\lambda_{\mathrm{faith}} = 0.5$, and $\lambda_{\mathrm{aux}} = 0.3$. All three classification terms, $\mathcal{L}_{\mathrm{cls}}$, $\mathcal{L}_{\mathrm{site}}$ and $\mathcal{L}_{\mathrm{aux}}$, are binary cross-entropy. Optimization uses decoupled weight decay \cite{loshchilov2019adamw}. The training uses gradient checkpoints and is conducted for a total of six epochs.

\subsection*{Evidence Branch and Re-injection}

Using the same backbone language model, adjustments are made by combining a low-rank adapter \cite{hu2022lora} with a rank of 16, a scaling factor of 32, and a dropout rate of 0.05. The vision tower remains frozen at this stage, so the two branches share a visual encoding. This branch generates item-level evidence with direction, severity, and time window, and answers localization questions in both directions. The evidence vector $e$ is the mean of the last-layer hidden states of the language model over the whole input sequence of frames and instruction, computed without generating text. It is projected to 256 dimensions by $W_e$ and concatenated with $h_v$ and with a 16-dimensional encoding $g(P_0)$ of the screening probability $P_0$ of the decision head. The concatenation passes through a fusion layer $f$, a $528 \to 256$ linear layer with layer normalization and GELU, and then a new decision head. The evidence is injected once.
\begin{equation}
\hat{y}=\sigma\! \big(w^{\top} f\big([\,h_v;\,W_e e;\,g(P_0)\,]\big)+b\big).
\end{equation}

\subsection*{Integration and Working Points}

All channels are fused before the decision head, so the deployed probability $\hat{y}$ is used directly. Therefore, no channel-level stacking is required after training. For test fold $k$, the threshold is taken from the positive rate $\pi_k$ of the training fold, rather than being fixed at half.
\begin{equation}
\tau_S=Q_{1-\pi_k}\big(\{\hat{y}_i : i\in S\}\big),
\end{equation}
Here, $Q_q$ represents the $q$th quantile and $S$ is a subsample of test fold $k$. The labels of the test fold do not affect the calculation of the working point.

\subsection*{Evaluation}

Each metric is calculated within each fold, and then the mean and standard deviation across the five folds are reported. Within each test fold, 200 random subsamples of half the fold are drawn without replacement, the threshold is set on each subsample with the rule above, and the six metrics are averaged over the 200 subsamples to give the value of that fold. For the concatenated out-of-fold predictions, the summary metrics are not reported because the probability magnitudes of different folds are not comparable.
Comparable video-based screening systems are commonly evaluated within a single cohort \cite{kojovic2021using,deng2024hear,natraj2024video,kim2025automated}, so we also report a stricter protocol. We randomly select sites from the 27 sites, combine them into 6 test sites, and hold them out in full, so that no recording from these sites enters training. The whole pipeline is retrained on the remaining sites, and cross-site performance is evaluated on the 6 held-out sites.

\section*{Discussion}

This study proposes ASDchat, a multimodal large language model that integrates video, audio, and dialogue. It has achieved stable results in ASD screening and behavioral subtype classification. The core design of ASDchat is an evidence-based reasoning framework, which connects automated screening and clinical accountability. Unlike black-box systems that only provide a raw probability score, ASDchat generates traceable, timestamped behavioral evidence and aligns it with standard clinical tools such as ADOS-2. This design addresses a central obstacle in clinical application, namely the distrust of clinicians and caregivers toward opaque AI systems in high-risk pediatric screening. Practitioners can directly view and verify the specific behavioral markers behind each judgment, which makes ASDchat more likely to gain clinical trust and turns automated screening into actionable pre-diagnostic interventions.

The vast majority of existing machine learning-based ASD screening models are designed for the binary classification task of distinguishing ASD from TD children \cite{lu2025autism}. However, these models inherently cannot distinguish ASD from other neurodevelopmental disorders, because the behavioral phenotypes of the latter often overlap with ASD. Previous studies have used electroencephalogram signals \cite{lira2026eeg}, eye-tracking data \cite{wei2024early}, and multimodal appearance features \cite{lu2025autism} to assess ASD against TD, and have systematically excluded other neurodevelopmental disorders. Even recent multimodal frameworks, such as video-audio neural network ensembles \cite{natraj2024video} and the vision-language model CARE-VL \cite{yoo2025care}, are limited to this binary discrimination. This narrow scope leaves a clinical translation gap. In reality, screening must address the high comorbidity and phenotypic overlap between ASD and intellectual disability, ADHD, and developmental language disorder \cite{mengi2022artificial}. ASDchat is different. We evaluated it both on distinguishing ASD from TD children and on the harder task of distinguishing ASD from other developmental disorders. Performance on the second task drops markedly, which reflects the difficulty encountered in actual clinical diagnosis and shows that screening tools must be evaluated against clinically realistic benchmarks that include comorbid and differential diagnostic conditions. To our knowledge, ASDchat is one of the few multimodal screening frameworks that systematically evaluates ASD against other developmental disorders, and it provides a more clinically meaningful evaluation that reflects the complexity of differential diagnosis.

In addition to binary screening, ASDchat further advances the field by enabling precise classification of the behavioral subtypes of ASD. Existing screening models only provide classification results and fail to reveal the heterogeneous phenotypic manifestations behind the autism spectrum. This heterogeneity is widely regarded as a major obstacle to personalized intervention planning, as individuals with ASD differ widely in social communication, restricted and repetitive behaviors, and cognitive functioning \cite{yuwattana2025machine}. Recent studies have attempted to achieve ASD subtyping using neuroimaging data such as functional connectivity maps, genomic and transcriptomic profiles, or parent-report questionnaires, but these methods typically require specialized equipment, invasive procedures, or extensive clinical resources. In contrast, we perform unsupervised clustering on seven behavioral dimensions computed statistically from the per-window annotations of the video recordings, and identify six reproducible ASD subtypes with different phenotypic profiles. This behavioral classification scheme requires no specialized equipment and no invasive procedures, is readily scalable, and supports early identification and the design of intervention plans. The model also provides intervention recommendations based on the subtype result. Each subtype is matched with strategies such as social-emotional training and structured communication support, which brings computational phenotyping into clinical work. This capability shifts screening from a population-level approach to personalized care based on subtypes.

\subsection*{Limitations and Future Work}

\begin{itemize}
    \item Dataset composition and generalizability: The evaluation of ASDchat is based on a multi-site dataset covering 27 research sites in three provinces in China. The geographical and ethnic coverage of the samples is limited. The male-to-female ratio in the training samples is approximately 4:1, which is consistent with the known sex distribution of individuals with ASD, but also limits the model's ability to learn behavioral patterns specific to females. The imbalance of samples leads to greater variance in the results for the female group and lower average performance, which indicates the need to construct a more balanced dataset and to evaluate the model separately by sex. Expanding the data collection scope to include people from different regions, cultural backgrounds, and socioeconomic levels can improve the generalizability of the model in a broader population.
    \item Categorical screening and diagnostic refinement: The current framework is mainly applicable to binary screening and cannot support detailed differential diagnosis. The distinction between ASD and other developmental disorders is difficult, which indicates that the model cannot fully capture the overlapping behavioral characteristics across diagnostic categories. In the future, multi-class classification tasks and richer diagnostic labels can be introduced, together with hierarchical classification strategies that further clarify the boundaries between ASD and related disorders. Using longitudinal behavioral data instead of cross-sectional data can also help the model capture how behavioral phenotypes change over development, and improve diagnostic accuracy.
    \item Integration with existing clinical workflows: When ASDchat is deployed in real clinical settings, workflow integration, data privacy, and regulatory compliance need to be addressed. The system relies on video collection, which raises privacy risks. These risks need to be controlled through secure storage, de-identification, and standardized informed consent procedures. Integrating ASDchat into routine pediatric screening, connecting it to electronic health records, and making it compatible with existing clinical assessment protocols require collaboration with healthcare providers and medical systems. Future work can adopt implementation science approaches to evaluate the usability and acceptability of the tool in different clinical settings, so that the system supports clinical work.
\end{itemize}

\section*{Conclusion}

This paper presents ASDchat, an evidence-based multimodal large language model for ASD screening. The model integrates video, audio, and dialogue within a unified framework, and outputs timestamped and traceable behavioral evidence through the evidence branch. On a 27-site dataset of 1,035 participants from China, ASDchat shows robust screening performance. This framework can identify six ASD behavioral subtypes and supports personalized intervention based on these subtypes. ASDchat provides a technical path for automated ASD screening, and alleviates the screening pressure on the medical system.

\section*{Data availability}

The dataset used in the current study was collected and deposited by the National Special Education Resource Center for Children with Autism in China. It is not publicly available because of privacy and confidentiality concerns, but is available from the corresponding author upon reasonable request.

\section*{Code availability}

The code used for model inference will be made available after publication. Code necessary for verification can be provided to editors and reviewers during the peer-review process.



\section*{Author Contributions}

Jun Chen and Qi Zhao conceived the idea, constructed the overall MLLM framework, and drafted the manuscript. Shuqin Cao, Yunqiang Lin, Chenglong Jia, Qiang Guo, and Xiaoyue Ma contributed to data collection and interpretation. Jun Chen, Guang Dai, Xiongtao Zhang, Mengmeng Wang, and Hui Wang contributed to model analysis. Yunliang Jiang and Wenguan Wang supervised the project.

\section*{Competing Interests}

The authors declare no competing interests.

\bibliographystyle{plain}  
\bibliography{sn-bibliography}  

@article{lyall2017changing,
  title={The changing epidemiology of autism spectrum disorders},
  author={Lyall, Kristen and Croen, Lisa and Daniels, Julie and Fallin, M Daniele and Ladd-Acosta, Christine and Lee, Brian K and Park, Bo Y and Snyder, Nathaniel W and Schendel, Diana and Volk, Heather and others},
  journal={Annual review of public health},
  volume={38},
  pages={81--102},
  year={2017},
  publisher={Annual Reviews}
}

@article{zeidan2022global,
  title={Global prevalence of autism: A systematic review update},
  author={Zeidan, Jinan and Fombonne, Eric and Scorah, Julie and Ibrahim, Alaa and Durkin, Maureen S and Saxena, Shekhar and Yusuf, Afiqah and Shih, Andy and Elsabbagh, Mayada},
  journal={Autism research},
  volume={15},
  number={5},
  pages={778--790},
  year={2022},
  publisher={Wiley Online Library}
}

@article{maenner2023prevalence,
  title={Prevalence and characteristics of autism spectrum disorder among children aged 8 years—Autism and Developmental Disabilities Monitoring Network, 11 sites, United States, 2020},
  author={Maenner, Matthew J},
  journal={MMWR. Surveillance summaries},
  volume={72},
  year={2023}
}

@article{lord2018autism,
  title={Autism spectrum disorder},
  author={Lord, Catherine and Elsabbagh, Mayada and Baird, Gillian and Veenstra-Vanderweele, Jeremy},
  journal={The lancet},
  volume={392},
  number={10146},
  pages={508--520},
  year={2018},
  publisher={Elsevier}
}

@article{dawson2010randomized,
  title={Randomized, controlled trial of an intervention for toddlers with autism: the Early Start Denver Model},
  author={Dawson, Geraldine and Rogers, Sally and Munson, Jeffrey and Smith, Milani and Winter, Jamie and Greenson, Jessica and Donaldson, Amy and Varley, Jennifer},
  journal={Pediatrics},
  volume={125},
  number={1},
  pages={e17--e23},
  year={2010},
  publisher={American Academy of Pediatrics}
}

@article{marin2016developmental,
  title={Developmental timing and critical windows for the treatment of psychiatric disorders},
  author={Mar{\'\i}n, Oscar},
  journal={Nature medicine},
  volume={22},
  number={11},
  pages={1229--1238},
  year={2016},
  publisher={Nature Publishing Group UK London}
}

@article{campbell2017use,
  title={Use of a digital modified checklist for autism in toddlers--revised with follow-up to improve quality of screening for autism},
  author={Campbell, Kathleen and Carpenter, Kimberly LH and Espinosa, Steven and Hashemi, Jordan and Qiu, Qiang and Tepper, Mariano and Calderbank, Robert and Sapiro, Guillermo and Egger, Helen L and Baker, Jeffrey P and others},
  journal={The Journal of Pediatrics},
  volume={183},
  pages={133--139},
  year={2017},
  publisher={Elsevier}
}

@article{sun2019autism,
  title={Autism prevalence in China is comparable to Western prevalence},
  author={Sun, Xiang and Allison, Carrie and Wei, Liping and Matthews, Fiona E and Auyeung, Bonnie and Wu, Yu Yu and Griffiths, Sian and Zhang, Jie and Baron-Cohen, Simon and Brayne, Carol},
  journal={Molecular autism},
  volume={10},
  number={1},
  pages={7},
  year={2019},
  publisher={Springer}
}

@article{zhou2020prevalence,
  title={Prevalence of autism spectrum disorder in China: a nationwide multi-center population-based study among children aged 6 to 12 years},
  author={Zhou, Hao and Xu, Xiu and Yan, Weili and Zou, Xiaobing and Wu, Lijie and Luo, Xuerong and Li, Tingyu and Huang, Yi and Guan, Hongyan and Chen, Xiang and others},
  journal={Neuroscience bulletin},
  volume={36},
  number={9},
  pages={961},
  year={2020}
}

@incollection{constantino2021social,
  title={Social responsiveness scale},
  author={Constantino, John N},
  booktitle={Encyclopedia of autism spectrum disorders},
  pages={4457--4467},
  year={2021},
  publisher={Springer}
}

@book{schopler2010childhood,
  title={The childhood autism rating scale, (CARS2): Manual},
  author={Schopler, Eric and Van Bourgondien, Marye E and Wellman, G Janette and Love, Steven R},
  year={2010},
  publisher={Western Psychological Services}
}

@article{lord2012autism,
  title={Autism diagnostic observation schedule--2nd edition (ADOS-2)},
  author={Lord, Catherine and Rutter, Michael and DiLavore, Pamel and Risi, Susan and Gotham, Katherine and Bishop, Somer and others},
  journal={Los Angeles, CA: Western Psychological Corporation},
  volume={284},
  pages={474--478},
  year={2012}
}

@article{wall2012use,
  title={Use of machine learning to shorten observation-based screening and diagnosis of autism},
  author={Wall, Dennis Paul and Kosmicki, Jack and Deluca, Todd F and Harstad, Elizabeth and Fusaro, Vincent Alfred},
  journal={Translational psychiatry},
  volume={2},
  number={4},
  pages={e100--e100},
  year={2012},
  publisher={Nature Publishing Group}
}

@article{insel2017digital,
  title={Digital phenotyping: technology for a new science of behavior},
  author={Insel, Thomas R},
  journal={Jama},
  volume={318},
  number={13},
  pages={1215--1216},
  year={2017}
}

@article{kojovic2021using,
  title={Using 2D video-based pose estimation for automated prediction of autism spectrum disorders in young children},
  author={Kojovic, Nada and Natraj, Shreyasvi and Mohanty, Sharada Prasanna and Maillart, Thomas and Schaer, Marie},
  journal={Scientific Reports},
  volume={11},
  number={1},
  pages={15069},
  year={2021},
  publisher={Nature Publishing Group UK London}
}

@article{chong2020detection,
  title={Detection of eye contact with deep neural networks is as accurate as human experts},
  author={Chong, Eunji and Clark-Whitney, Elysha and Southerland, Audrey and Stubbs, Elizabeth and Miller, Chanel and Ajodan, Eliana L and Silverman, Melanie R and Lord, Catherine and Rozga, Agata and Jones, Rebecca M and others},
  journal={Nature communications},
  volume={11},
  number={1},
  pages={6386},
  year={2020},
  publisher={Nature Publishing Group UK London}
}

@article{jabbar2026deep,
  title={Deep learning based approach for Behavior classification in diagnoses of Autism Spectrum Disorder using naturalistic videos},
  author={Jabbar, Usama and Waseem Iqbal, Muhammad and Nechifor, Alexandru and Abaker, Mohammed and Khairalseed, Mohammed Ahmed and Antohi, Valentin Marian and Fortea, Costinela and Stefanescu, Catalin Aurelian},
  journal={Frontiers in Computational Neuroscience},
  volume={20},
  pages={1626315},
  year={2026},
  publisher={Frontiers Media SA}
}

@article{khan2025ws,
  title={WS-BiTM: Integrating White Shark Optimization with Bi-LSTM for enhanced autism spectrum disorder diagnosis},
  author={Khan, Kainat and Katarya, Rahul},
  journal={Journal of Neuroscience Methods},
  volume={413},
  pages={110319},
  year={2025},
  publisher={Elsevier}
}

@inproceedings{boluk2025gaze,
  title={Gaze Analysis of Children with Autism During Robot-Assisted Therapy},
  author={Boluk, Nursena and Kose, Hatice},
  booktitle={Proceedings of the 2025 Symposium on Eye Tracking Research and Applications},
  pages={1--6},
  year={2025}
}

@article{nafisah2025deep,
  title={Deep learning-based feature selection for detection of autism spectrum disorder},
  author={Nafisah, Ibrahim and Mahmoud, Nermine and Ewees, Ahmed A and Khattap, Mohamed G and Dahou, Abdelghani and Alghamdi, Safar M and Fares, Ibrahim A and Azmi Al-Betar, Mohammed and Abd Elaziz, Mohamed},
  journal={Frontiers in Artificial Intelligence},
  volume={8},
  pages={1594372},
  year={2025},
  publisher={Frontiers Media SA}
}

@inproceedings{kommineni2025can,
  title={Can Multimodal Foundation Models Help Analyze Child-Inclusive Autism Diagnostic Videos?},
  author={Kommineni, Aditya and Bose, Digbalay and Feng, Tiantian and Kim, So Hyun and Tager-Flusberg, Helen and Bishop, Somer and Lord, Catherine and Kadiri, Sudarsana and Narayanan, Shrikanth},
  booktitle={Proc. Interspeech 2025},
  pages={3050--3054},
  year={2025}
}

@inproceedings{yoo2025care,
  title={CARE-VL: A Domain-Specialized Vision-Language Model for Early ASD Screening},
  author={Yoo, Cheol-Hwan and Yoo, Jang-Hee and Jang, Jaeyoon},
  booktitle={International Conference on Medical Image Computing and Computer-Assisted Intervention},
  pages={57--66},
  year={2025},
  organization={Springer}
}

@inproceedings{zhong2025multi,
  title={Multi-modal progressive fusion for ASD screening using smartphone video},
  author={Zhong, Wenqi and Li, Bohan and Xia, Chen and Li, Kuan and Zhang, Dingwen},
  booktitle={International Conference on Medical Image Computing and Computer-Assisted Intervention},
  pages={393--403},
  year={2025},
  organization={Springer}
}

@article{kim2025automated,
  title={Automated AI based identification of autism spectrum disorder from home videos},
  author={Kim, Dong Yeong and Do, Ryemi and Shin, Youmin and Sim, Hewoen and Kim, Hanna and Cho, Sungchul and Lee, Geonhee and Park, Seyeon and Jang, Boa and Lim, Hyojeong and others},
  journal={npj Digital Medicine},
  volume={8},
  number={1},
  pages={607},
  year={2025},
  publisher={Nature Publishing Group UK London}
}

@article{deng2024hear,
  title={Hear me, see me, understand me: Audio-visual autism behavior recognition},
  author={Deng, Shijian and Kosloski, Erin E and Patel, Siddhi and Barnett, Zeke A and Nan, Yiyang and Kaplan, Alexander and Aarukapalli, Sisira and Doan, William T and Wang, Matthew and Singh, Harsh and others},
  journal={IEEE Transactions on Multimedia},
  volume={27},
  pages={2335--2346},
  year={2024},
  publisher={IEEE}
}

@article{natraj2024video,
  title={Video-audio neural network ensemble for comprehensive screening of autism spectrum disorder in young children},
  author={Natraj, Shreyasvi and Kojovic, Nada and Maillart, Thomas and Schaer, Marie},
  journal={Plos one},
  volume={19},
  number={10},
  pages={e0308388},
  year={2024},
  publisher={Public Library of Science San Francisco, CA USA}
}

@article{marey2024explainability,
  title={Explainability, transparency and black box challenges of AI in radiology: impact on patient care in cardiovascular radiology},
  author={Marey, Ahmed and Arjmand, Parisa and Alerab, Ameerh Dana Sabe and Eslami, Mohammad Javad and Saad, Abdelrahman M and Sanchez, Nicole and Umair, Muhammad},
  journal={Egyptian Journal of Radiology and Nuclear Medicine},
  volume={55},
  number={1},
  pages={183},
  year={2024},
  publisher={Springer}
}

@article{tang2014loss,
  title={Loss of mTOR-dependent macroautophagy causes autistic-like synaptic pruning deficits},
  author={Tang, Guomei and Gudsnuk, Kathryn and Kuo, Sheng-Han and Cotrina, Marisa L and Rosoklija, Gorazd and Sosunov, Alexander and Sonders, Mark S and Kanter, Ellen and Castagna, Candace and Yamamoto, Ai and others},
  journal={Neuron},
  volume={83},
  number={5},
  pages={1131--1143},
  year={2014},
  publisher={Elsevier}
}

@article{fyfe2026time,
  title={Time trends in the male to female ratio for autism incidence: population based, prospectively collected, birth cohort study},
  author={Fyfe, Caroline and Winell, Henric and Dougherty, Joseph and Gutmann, David H and Kolevzon, Alexander and Marrus, Natasha and Tedroff, Kristina and Turner, Tychele N and Weiss, Lauren A and Yip, Benjamin HK and others},
  journal={bmj},
  volume={392},
  year={2026},
  publisher={British Medical Journal Publishing Group}
}

@article{bruni2014test,
  title={Test review: Social responsiveness scale--Second edition (SRS-2)},
  author={Bruni, Teryn P},
  journal={Journal of Psychoeducational Assessment},
  volume={32},
  number={4},
  pages={365--369},
  year={2014},
  publisher={Sage Publications Sage CA: Los Angeles, CA}
}

@article{robins2014validation,
  title={Validation of the modified checklist for autism in toddlers, revised with follow-up (M-CHAT-R/F)},
  author={Robins, Diana L and Casagrande, Kar{\'\i}s and Barton, Marianne and Chen, Chi-Ming A and Dumont-Mathieu, Thyde and Fein, Deborah},
  journal={Pediatrics},
  volume={133},
  number={1},
  pages={37--45},
  year={2014},
  publisher={American Academy of Pediatrics Elk Grove Village, IL, USA}
}

@article{pagani2026autism,
  title={Autism subtypes identified using cross-species functional connectivity analyses},
  author={Pagani, Marco and Zerbi, Valerio and Gini, Silvia and Alvino, Filomena Grazia and Banerjee, Abhishek and Barberis, Andrea and Basson, M Albert and Bozzi, Yuri and Galbusera, Alberto and Ellegood, Jacob and others},
  journal={Nature Neuroscience},
  pages={1--12},
  year={2026},
  publisher={Nature Publishing Group US New York}
}

@article{shnitko2026parsing,
  title={Parsing autism spectrum heterogeneity through fMRI: Autism},
  author={Shnitko, Tatiana A and Lin, Shih-Che Alex and Shih, Yen-Yu Ian},
  journal={Nature neuroscience},
  pages={1--4},
  year={2026},
  publisher={Nature Publishing Group US New York}
}

@article{buch2023molecular,
  title={Molecular and network-level mechanisms explaining individual differences in autism spectrum disorder},
  author={Buch, Amanda M and V{\'e}rtes, Petra E and Seidlitz, Jakob and Kim, So Hyun and Grosenick, Logan and Liston, Conor},
  journal={Nature neuroscience},
  volume={26},
  number={4},
  pages={650--663},
  year={2023},
  publisher={Nature Publishing Group US New York}
}

@article{litman2025decomposition,
  title={Decomposition of phenotypic heterogeneity in autism reveals underlying genetic programs},
  author={Litman, Aviya and Sauerwald, Natalie and Green Snyder, Leeanne and Foss-Feig, Jennifer and Park, Christopher Y and Hao, Yun and Dinstein, Ilan and Theesfeld, Chandra L and Troyanskaya, Olga G},
  journal={Nature Genetics},
  volume={57},
  number={7},
  pages={1611--1619},
  year={2025},
  publisher={Nature Publishing Group US New York}
}

@inproceedings{he2017maskrcnn,
  title={Mask {R-CNN}},
  author={He, Kaiming and Gkioxari, Georgia and Doll{\'a}r, Piotr and Girshick, Ross},
  booktitle={Proceedings of the IEEE International Conference on Computer Vision},
  pages={2961--2969}, year={2017}}

@inproceedings{cho2014gru,
  title={Learning Phrase Representations using {RNN} Encoder--Decoder for Statistical Machine Translation},
  author={Cho, Kyunghyun and van Merri{\"e}nboer, Bart and Gulcehre, Caglar and Bahdanau, Dzmitry
          and Bougares, Fethi and Schwenk, Holger and Bengio, Yoshua},
  booktitle={Proceedings of the 2014 Conference on Empirical Methods in Natural Language Processing},
  pages={1724--1734}, year={2014}}

@inproceedings{ganin2015grl,
  title={Unsupervised Domain Adaptation by Backpropagation},
  author={Ganin, Yaroslav and Lempitsky, Victor},
  booktitle={Proceedings of the 32nd International Conference on Machine Learning},
  pages={1180--1189}, year={2015}}

@inproceedings{gao2023funasr,
  title={{FunASR}: A Fundamental End-to-End Speech Recognition Toolkit},
  author={Gao, Zhifu and Li, Zerui and Wang, Jiaming and Luo, Haoneng and Shi, Xian and Chen, Mengzhe
          and Li, Yabin and Zuo, Lingyun and Du, Zhihao and Xiao, Zhangyu and Zhang, Shiliang},
  booktitle={Proceedings of INTERSPEECH}, year={2023}}

@inproceedings{bredin2020pyannote,
  title={pyannote.audio: Neural Building Blocks for Speaker Diarization},
  author={Bredin, Herv{\'e} and Yin, Ruiqing and Coria, Juan Manuel and Gelly, Gregory
          and Korshunov, Pavel and Lavechin, Marvin and Fustes, Diego and Titeux, Hadrien
          and Bouaziz, Wassim and Gill, Marie-Philippe},
  booktitle={IEEE International Conference on Acoustics, Speech and Signal Processing},
  pages={7124--7128}, year={2020}}

@book{lord2012ados2,
  title={Autism Diagnostic Observation Schedule, Second Edition {(ADOS-2)} Manual},
  author={Lord, Catherine and Rutter, Michael and DiLavore, Pamela C. and Risi, Susan
          and Gotham, Katherine and Bishop, Somer L.},
  publisher={Western Psychological Services}, address={Torrance, CA}, year={2012}}

@article{qwen3vl2025,
  title={{Qwen3-VL} Technical Report},
  author={{Qwen Team}},
  journal={arXiv preprint arXiv:2511.21631}, year={2025}}

@inproceedings{tran2018r21d,
  title={A Closer Look at Spatiotemporal Convolutions for Action Recognition},
  author={Tran, Du and Wang, Heng and Torresani, Lorenzo and Ray, Jamie and LeCun, Yann and Paluri, Manohar},
  booktitle={Proceedings of the IEEE Conference on Computer Vision and Pattern Recognition},
  pages={6450--6459}, year={2018}}

@article{eyben2016gemaps,
  title={The {Geneva} Minimalistic Acoustic Parameter Set {(GeMAPS)} for Voice Research and Affective Computing},
  author={Eyben, Florian and Scherer, Klaus R. and Schuller, Bj{\"o}rn W. and Sundberg, Johan
          and Andr{\'e}, Elisabeth and Busso, Carlos and Devillers, Laurence Y. and Epps, Julien
          and Laukka, Petri and Narayanan, Shrikanth S. and Truong, Khiet P.},
  journal={IEEE Transactions on Affective Computing}, volume={7}, number={2},
  pages={190--202}, year={2016}}

@inproceedings{hu2022lora,
  title={{LoRA}: Low-Rank Adaptation of Large Language Models},
  author={Hu, Edward J. and Shen, Yelong and Wallis, Phillip and Allen-Zhu, Zeyuan and Li, Yuanzhi
          and Wang, Shean and Wang, Lu and Chen, Weizhu},
  booktitle={International Conference on Learning Representations}, year={2022}}

@article{kay2017kinetics,
  title={The {Kinetics} Human Action Video Dataset},
  author={Kay, Will and Carreira, Joao and Simonyan, Karen and Zhang, Brian and Hillier, Chloe
          and Vijayanarasimhan, Sudheendra and Viola, Fabio and Green, Tim and Back, Trevor
          and Natsev, Paul and Suleyman, Mustafa and Zisserman, Andrew},
  journal={arXiv preprint arXiv:1705.06950}, year={2017}}

@inproceedings{loshchilov2019adamw,
  title={Decoupled Weight Decay Regularization},
  author={Loshchilov, Ilya and Hutter, Frank},
  booktitle={International Conference on Learning Representations}, year={2019}}

@article{rousseeuw1987silhouettes,
  title={Silhouettes: A Graphical Aid to the Interpretation and Validation of Cluster Analysis},
  author={Rousseeuw, Peter J.},
  journal={Journal of Computational and Applied Mathematics}, volume={20},
  pages={53--65}, year={1987}}

@inproceedings{lira2026eeg,
  title={EEG Topographic Mapping and Deep Transfer Learning for Early Detection of Autism Spectrum Disorder},
  author={Lira, Maria Eduarda and Fonseca, Fl{\'a}vio Secco and Sayonara, Adrielly and Viana, Arianne and Cordeiro, Cec{\'\i}lia and Santana, Ma{\'\i}ra and Gomes, Juliana C and Dos Santos, Wellington P},
  booktitle={2026 9th International Conference on Artificial Intelligence and Big Data (ICAIBD)},
  pages={661--666},
  year={2026},
  organization={IEEE}
}

@article{wei2024early,
  title={Early identification of autism spectrum disorder based on machine learning with eye-tracking data},
  author={Wei, Qiuhong and Dong, Wenxin and Yu, Dongchuan and Wang, Ke and Yang, Ting and Xiao, Yuanjie and Long, Dan and Xiong, Haiyi and Chen, Jie and Xu, Ximing and others},
  journal={Journal of affective disorders},
  volume={358},
  pages={326--334},
  year={2024},
  publisher={Elsevier}
}

@article{lu2025autism,
  title={Autism screening for children based on appearance features across multiple paradigms},
  author={Lu, Xiaofeng and Yue, Siyao and Li, Chaozhen and Yang, Xia and Wang, Yulin and Liu, Zhi},
  journal={Displays},
  volume={88},
  pages={103049},
  year={2025},
  publisher={Elsevier}
}

@article{mengi2022artificial,
  title={Artificial intelligence based techniques for the detection of socio-behavioral disorders: a systematic review},
  author={Mengi, Mehak and Malhotra, Deepti},
  journal={Archives of Computational Methods in Engineering},
  volume={29},
  number={5},
  pages={2811--2855},
  year={2022},
  publisher={Springer}
}

@article{yuwattana2025machine,
  title={Machine learning of clinical phenotypes facilitates autism screening and identifies novel subgroups with distinct transcriptomic profiles},
  author={Yuwattana, Wasana and Saeliw, Thanit and van Erp, Marlieke Lisanne and Poolcharoen, Chayanit and Kanlayaprasit, Songphon and Trairatvorakul, Pon and Chonchaiya, Weerasak and Hu, Valerie W and Sarachana, Tewarit},
  journal={Scientific Reports},
  volume={15},
  number={1},
  pages={11712},
  year={2025},
  publisher={Nature Publishing Group UK London}
}

@article{moor2023foundation,
  title={Foundation models for generalist medical artificial intelligence},
  author={Moor, Michael and Banerjee, Oishi and Abad, Zahra Shakeri Hossein and Krumholz, Harlan M and Leskovec, Jure and Topol, Eric J and Rajpurkar, Pranav},
  journal={Nature},
  volume={616},
  number={7956},
  pages={259--265},
  year={2023},
  publisher={Nature Publishing Group UK London}
}

@article{singhal2023large,
  title={Large language models encode clinical knowledge},
  author={Singhal, Karan and Azizi, Shekoofeh and Tu, Tao and Mahdavi, S Sara and Wei, Jason and Chung, Hyung Won and Scales, Nathan and Tanwani, Ajay and Cole-Lewis, Heather and Pfohl, Stephen and others},
  journal={Nature},
  volume={620},
  pages={172--180},
  year={2023},
  doi={10.1038/s41586-023-06291-2},
  publisher={Nature Publishing Group UK London}
}

@article{fortin2018harmonization,
  title={Harmonization of cortical thickness measurements across scanners and sites},
  author={Fortin, Jean-Philippe and Cullen, Nicholas and Sheline, Yvette I and Taylor, Warren D and Aselcioglu, Irem and Cook, Philip A and Adams, Phil and Cooper, Crystal and Fava, Maurizio and McGrath, Patrick J and others},
  journal={Neuroimage},
  volume={167},
  pages={104--120},
  year={2018},
  publisher={Elsevier}
}

@article{washington2023review,
  title={A Review of and Roadmap for Data Science and Machine Learning for the Neuropsychiatric Phenotype of Autism},
  author={Washington, Peter and Wall, Dennis P},
  journal={Annual Review of Biomedical Data Science},
  volume={6},
  pages={211--228},
  year={2023},
  doi={10.1146/annurev-biodatasci-020722-125454},
  publisher={Annual Reviews}
}

@article{zwaigenbaum2015early,
  title={Early identification of autism spectrum disorder: recommendations for practice and research},
  author={Zwaigenbaum, Lonnie and Bauman, Margaret L and Stone, Wendy L and Yirmiya, Nurit and Estes, Annette and Hansen, Robin L and McPartland, James C and Natowicz, Marvin R and Choueiri, Roula and Fein, Deborah and others},
  journal={Pediatrics},
  volume={136},
  number={Supplement\_1},
  pages={S10--S40},
  year={2015},
  publisher={American Academy of Pediatrics Elk Grove Village, IL, USA}
}

%

\end{document}